\documentclass[sigconf]{acmart}
    
\usepackage{multirow}
\usepackage{graphicx}
\usepackage{float}
\usepackage{siunitx}
\usepackage{balance}
\newcolumntype{j}{S[table-format=3.]}
\usepackage{xcolor}

\copyrightyear{2023} 
\acmYear{2023} 
\setcopyright{acmlicensed}\acmConference[SOICT 2023]{The 12th International Symposium on Information and Communication Technology}{December 7--8, 2023}{Ho Chi Minh, Vietnam}
\acmBooktitle{The 12th International Symposium on Information and Communication Technology (SOICT 2023), December 7--8, 2023, Ho Chi Minh, Vietnam}
\acmPrice{15.00}
\acmDOI{10.1145/3628797.3628837}
\acmISBN{979-8-4007-0891-6/23/12}

\begin{document}

\title{Evaluating the Symbol Binding Ability of Large Language Models for Multiple-Choice Questions in Vietnamese General Education}

\renewcommand{\shorttitle}{Evaluating the Symbol Binding Ability of Large Language Models \\ for Multiple-Choice Questions in Vietnamese General Education}

\author{Duc-Vu Nguyen}
\authornote{Both authors contributed equally to this research.}
\affiliation{%
  \institution{University of Information Technology}
  \city{Ho Chi Minh City}
  \country{Vietnam}
}
\affiliation{%
  \institution{Vietnam National University}
  \city{Ho Chi Minh City}
  \country{Vietnam}
}
\email{vund@uit.edu.vn}

\author{Quoc-Nam Nguyen}
\authornotemark[1]
\affiliation{%
  \institution{University of Information Technology}
  \city{Ho Chi Minh City}
  \country{Vietnam}
}
\affiliation{%
  \institution{Vietnam National University}
  \city{Ho Chi Minh City}
  \country{Vietnam}
}
\email{20520644@gm.uit.edu.vn}

\begin{abstract}
\sloppy In this paper, we evaluate the ability of large language models (LLMs) to perform multiple choice symbol binding (MCSB) for multiple choice question answering (MCQA) tasks in zero-shot, one-shot, and few-shot settings. We focus on Vietnamese, with fewer challenging MCQA datasets than in English. The two existing datasets, ViMMRC 1.0 and ViMMRC 2.0, focus on literature. Recent research in Vietnamese natural language processing (NLP) has focused on the Vietnamese National High School Graduation Examination (VNHSGE) from 2019 to 2023 to evaluate ChatGPT. However, these studies have mainly focused on how ChatGPT solves the VNHSGE step by step. We aim to create a novel and high-quality dataset by providing structured guidelines for typing LaTeX formulas for mathematics, physics, chemistry, and biology. This dataset can be used to evaluate the MCSB ability of LLMs and smaller language models (LMs) because it is typed in a strict LaTeX style. We determine the most probable character answer (A, B, C, or D) based on context, instead of finding the answer step by step as in previous Vietnamese works. This reduces computational costs and accelerates the evaluation of LLMs. Our evaluation of six well-known LLMs, namely BLOOMZ-7.1B-MT, LLaMA-2-7B, LLaMA-2-70B, GPT-3, GPT-3.5, and GPT-4.0, on the ViMMRC 1.0 and ViMMRC 2.0 benchmarks and our proposed dataset shows promising results on the MCSB ability of LLMs for Vietnamese. The dataset is available\footnote{\url{https://huggingface.co/datasets/uitnlp/ViGEText_17to23}} for research purposes only.
\end{abstract}



\maketitle

\section{Introduction}

\sloppy Large language models (LLMs) have become instrumental in a wide array of natural language processing (NLP) \cite{brown2020language,muennighoff-etal-2023-crosslingual,touvron2023llama,OpenAI2023GPT4TR}. In the era of AI-driven advancements, the capability of LLMs to tackle complex challenges continues to be a subject of intense research and evaluation. One such challenge is the realm of multiple choice questing answering (MCQA) task, where LLMs are tasked with understanding contextual information and selecting the most appropriate answer from a set of choices. In this paper, we delve into the essential domain of multiple choice symbol binding (MCSB) \cite{robinson2023leveraging} in MCQA, aiming to shed light on LLMs' proficiency when faced with this intricate task.

While LLMs have demonstrated remarkable proficiency in various NLP tasks, the Vietnamese language presents unique challenges and opportunities. Unlike English, Vietnamese has limited challenging MCQA datasets available for research purposes. Existing datasets, such as ViMMRC 1.0 \cite{kietnv2020} and ViMMRC 2.0 \cite{sonlt2023}, primarily focus on literary contexts, leaving a substantial gap in assessing LLMs' capabilities across diverse domains.

In recent Vietnamese NLP research, the evaluation of models has largely centered on their ability to solve questions from the Vietnamese National High School Graduation Examination (VNHSGE) between 2019 and 2023. However, these studies predominantly analyze the step-by-step problem-solving process rather than focusing on the broader capacity of models for MCQA in Vietnamese.

Recognizing the need for a comprehensive dataset that encompasses a wide range of subjects and promotes the evaluation of LLMs' MCSB abilities, we have created a novel, high-quality dataset. This dataset includes structured guidelines for typing LaTeX formulas in Mathematics, Physics, Chemistry, and Biology. By enforcing a strict LaTeX formatting style, we aim to provide a standardized and meticulous evaluation environment that can be utilized not only for assessing LLMs but also for evaluating smaller language models (LMs).

Within the confines of this study, our primary objective is to predict the correct answer character (A, B, C, or D) for a given question, anchored in its contextual framework. To ensure a comprehensive evaluation, we assess the performance of six well-recognized LLMs: BLOOMZ-7.1B-MT, LLaMA-2-7B, LLaMA-2-70B, GPT-3, GPT-3.5, and GPT-4.0. Our evaluation encompasses the ViMMRC 1.0 and ViMMRC 2.0 benchmarks alongside our novel dataset. The outcomes of this exhaustive analysis provide invaluable insights into the MCSB capabilities of LLMs in the Vietnamese language, poised to influence future research and development endeavors in this domain. These findings pave the way for harnessing the full potential of language models in addressing the scarcity of challenging MCQA datasets in Vietnamese and refining their proficiency in specialized domains. 

Our contributions are summarized as follows:
\begin{enumerate}
    \item We presented a novel, high-quality dataset with structured guidelines for typing LaTeX formulas in Mathematics, Physics, Chemistry, and Biology.
    \item We conducted experiments on the symbol binding ability of LLMs for multiple-choice questions in the context of Vietnamese General Education. Our comprehensive evaluation includes six prominent LLMs, namely BLOOMZ-7.1B-MT, LLaMA-2-7B, LLaMA-2-70B, GPT-3, GPT-3.5, and GPT-4.0.
    \item Extensive analysis and discussion are made to figure out in-depth how LLMs impact Vietnamese multiple-choice questions on examinations and explore the implications of LLMs in education.
\end{enumerate}

\section{Related Work}

Comprehending and manipulating symbols within language is essential for effectively responding to multiple-choice questions.

In their work, \citeauthor{lai-etal-2017-race} \cite{lai-etal-2017-race} introduced RACE, a novel dataset created to assess methods in the field of reading comprehension. This dataset, comprised of nearly 28,000 passages and approximately 100,000 questions developed by English instructors, was collected from English exams taken by Chinese students aged 12 to 18 in middle and high schools. It encompasses a wide range of topics deliberately selected to evaluate students' abilities in comprehension and reasoning.

\citeauthor{hendrycksmath2021} \cite{hendrycksmath2021} presented MATH, a fresh dataset featuring 12,500 challenging mathematical problems designed for competitive assessments. Each problem in MATH includes a comprehensive, step-by-step solution, providing valuable resources for training models to generate answer derivations and explanations.

In the domain of science, \citeauthor{lu2022learn} \cite{lu2022learn} introduced SCIENCEQA, a new benchmark comprising approximately 21,000 multimodal multiple-choice questions spanning various science topics. The dataset also includes annotations for answers, corresponding lectures, and explanations.

\citeauthor{lewis-etal-2020-mlqa} \cite{lewis-etal-2020-mlqa} introduced MLQA, a cross-lingual extractive question-answering benchmark. MLQA consists of QA instances in seven languages: English, Arabic, German, Spanish, Hindi, Vietnamese, and Simplified Chinese. It encompasses over 12,000 instances in English and 5,000 in each language, each with parallel versions in an average of four languages.

\citeauthor{dao2023evaluation} \cite{dao2023evaluation} assessed ChatGPT (Feb 13 Version), a large language model, to evaluate its performance in addressing English test questions derived from the Vietnamese National High School Graduation Exam spanning the years 2019 to 2023. The findings of the study's analysis revealed that ChatGPT achieved an average accuracy rate of 40 correct responses out of 50 questions, corresponding to a score of 7.92 on the 10-point scale commonly employed in Vietnam. Notably, the accuracy of ChatGPT's answers remained consistent across varying levels of question difficulty, highlighting the model's proficiency in this particular task.

\section{Datasets}\label{dataset}
This section presents the datasets used to evaluate the symbol binding ability of Large Language Models.

\subsection{ViMMRC 1.0}\label{vimmrc1}
\citeauthor{kietnv2020} assembled a dataset \textbf{ViMMRC 1.0} with 2,783 sets of multiple-choice questions and their corresponding answers. These questions are drawn from 417 Vietnamese texts typically utilized in reading comprehension instruction for elementary school students.

\subsection{ViMMRC 2.0}
\textbf{ViMMRC 2.0} is introduced by \citeauthor{sonlt2023} to expand the earlier ViMMRC 1.0 (described in Section~\ref{vimmrc1}) dataset designed for multiple-choice reading comprehension in Vietnamese textbooks. \textbf{ViMMRC 2.0} comprises 699 reading passages, including prose and poems, and 5,273 questions. Unlike the previous version, this dataset does not constrain the questions to have fixed four options. Additionally, the questions in this new dataset are designed to be more challenging, requiring models to thoroughly comprehend the entire context of the reading passage, question, and the content of each available choice to extract the answers correctly.

\subsection{Our Proposed Dataset: ViGEText\_17to23}
Our proposed dataset has been meticulously assembled through web crawling from publicly accessible internet sources. Distinct from the prior study conducted by \citeauthor{dao2023vnhsge} \cite{dao2023vnhsge} between 2019 and 2023, our objective was to cover the entire scope of the Vietnamese General Education Examination spanning from 2017 to 2023. This comprehensive approach included the challenging examinations of the years 2017 and 2018, which have been significant for nearly all Vietnamese students in recent years. It is important to highlight that the exact and unquestionably correct answers have been exclusively obtained from the Vietnamese Ministry of Education. This approach was taken to uphold the highest standards of accuracy and reliability in our dataset. Table \ref{tab:summarydataset} presents the statistics for each subject from 2017 to 2023.

In \textbf{Mathematics}, \textbf{Physics}, \textbf{Biology}, and \textbf{Chemistry}, standardization necessitates the translation of geometry into a geometric language. However, implementing this update is time-consuming, thus prompting us to defer it to future works. Additionally, regarding the subject of \textbf{Geography}, statements containing temporal elements may not hold true (always accurate) in the near or distant future. Consequently, we have removed these statements to ensure the dataset remains reusable and subject to long-term evaluation.

\begin{table*}[t]
\centering
\resizebox{\textwidth}{!}{%
\begin{tabular}{|ccc||j|j|j|j|j|j|j||j|}
\hline
\multicolumn{1}{|c|}{\textbf{Year}} & \multicolumn{2}{c||}{\textbf{Type of Test}} & \textbf{Mathematics} & \textbf{Physics} & \textbf{Chemistry} & \textbf{Biology} & \textbf{History} & \textbf{Geography} & \textbf{\begin{tabular}[c]{@{}c@{}}Civic\\ Education\end{tabular}} & \textbf{Total} \\ \hline\hline
\multicolumn{1}{|c|}{\textbf{2017}} & \multicolumn{2}{c||}{\textbf{Actual}} & 45 & 37 & 37 & 35 & 40 & 28 & 40 & 262 \\ \hline
\multicolumn{1}{|c|}{\multirow{2}{*}{\textbf{2018}}} & \multicolumn{2}{c||}{\textbf{Sample}} & 39 & 34 & 37 & 37 & 40 & 24 & 40 & 251 \\ \cline{2-11} 
\multicolumn{1}{|c|}{} & \multicolumn{2}{c||}{\textbf{Actual}} & 43 & 33 & 37 & 39 & 40 & 18 & 40 & 250 \\ \hline
\multicolumn{1}{|c|}{\multirow{2}{*}{\textbf{2019}}} & \multicolumn{2}{c||}{\textbf{Sample}} & 36 & 35 & 38 & 39 & 40 & 22 & 40 & 250 \\ \cline{2-11} 
\multicolumn{1}{|c|}{} & \multicolumn{2}{c||}{\textbf{Actual}} & 36 & 36 & 38 & 32 & 40 & 20 & 40 & 242 \\ \hline
\multicolumn{1}{|c|}{\multirow{3}{*}{\textbf{2020}}} & \multicolumn{2}{c||}{\textbf{Sample}} & 37 & 35 & 39 & 39 & 40 & 21 & 40 & 251 \\ \cline{2-11} 
\multicolumn{1}{|c|}{} & \multicolumn{1}{c|}{\multirow{2}{*}{\textbf{Actual}}} & \textbf{Round 1} & 40 & 35 & 35 & 38 & 40 & 18 & 40 & 251 \\ \cline{3-11} 
\multicolumn{1}{|c|}{} & \multicolumn{1}{c|}{} & \textbf{Round 2} & 41 & 35 & 40 & 38 & 40 & 18 & 40 & 252 \\ \hline
\multicolumn{1}{|c|}{\multirow{3}{*}{\textbf{2021}}} & \multicolumn{2}{c||}{\textbf{Sample}} & 39 & 36 & 40 & 36 & 40 & 13 & 40 & 244 \\ \cline{2-11} 
\multicolumn{1}{|c|}{} & \multicolumn{1}{c|}{\multirow{2}{*}{\textbf{Actual}}} & \textbf{Round 1} & 42 & 35 & 40 & 35 & 40 & 13 & 40 & 245 \\ \cline{3-11} 
\multicolumn{1}{|c|}{} & \multicolumn{1}{c|}{} & \textbf{Round 2} & 41 & 35 & 40 & 37 & 40 & 14 & 40 & 247 \\ \hline
\multicolumn{1}{|c|}{\multirow{2}{*}{\textbf{2022}}} & \multicolumn{2}{c||}{\textbf{Sample}} & 43 & 35 & 40 & 36 & 40 & 11 & 40 & 245 \\ \cline{2-11} 
\multicolumn{1}{|c|}{} & \multicolumn{2}{c||}{\textbf{Actual}} & 42 & 36 & 39 & 37 & 40 & 13 & 40 & 247 \\ \hline
\multicolumn{1}{|c|}{\multirow{2}{*}{\textbf{2023}}} & \multicolumn{2}{c||}{\textbf{Sample}} & 41 & 36 & 38 & 34 & 40 & 14 & 40 & 243 \\ \cline{2-11} 
\multicolumn{1}{|c|}{} & \multicolumn{2}{c||}{\textbf{Actual}} & 41 & 37 & 38 & 33 & 40 & 13 & 40 & 242 \\ \hline\hline
\multicolumn{3}{|c||}{\textbf{Total}} & 606 & 530 & 581 & 545 & 600 & 260 & 600 & 3722 \\ \hline
\end{tabular}%
}
\caption{Our proposed dataset statistics. In light of COVID-19, the Vietnamese Ministry of Education adopted a regional two-round exam schedule for 2020 and 2021 as a precautionary measure.}
\label{tab:summarydataset}
\end{table*}

Our primary objective is establishing a comprehensive and high-caliber dataset from Vietnamese General Education. To achieve this, we are dedicated to supplying meticulously structured guidelines tailored to accurately input LaTeX formulas in mathematics, physics, chemistry, and biology. The overarching rationale behind this ambitious endeavor is to enhance the symbolic mathematics field significantly. Symbolic mathematics, often reliant on LaTeX notation for precision and versatility, plays a pivotal role in various scientific disciplines. This dataset-creation initiative is motivated by the pressing need to address the challenges researchers, educators, and students face when working with mathematical expressions, equations, and notations. 

For more detail, our proposed dataset was meticulously developed following the formatting standards of the \textbf{MathJax}\footnote{\url{https://www.mathjax.org/}} library in LaTeX. \textbf{With the goal of representing mathematical formulas according to strict rules, ensuring fair inference across all contexts, and enabling easy parsing in further research, even for small language models}, our LaTeX typing guidelines dictate that mathematical formulas must be written without spaces (except when writing chemical equations, as demonstrated in the chemical examples in Appendix~\ref{sampledata}), and curly brackets should only be used when necessary. After typing the formula, ensure it is displayed exactly as it appears in the original image of the authentic exam paper. Furthermore, our approach incorporated \textcolor{orange}{``\textit{\textbackslash{}ce}''} to accurately represent chemical elements, and \textcolor{blue}{``\textit{\textbackslash{}pu}''} effectively denotes measurement units (both are included in \textbf{mhchem}\footnote{\url{https://docs.mathjax.org/en/latest/input/tex/extensions/mhchem.html}} extensions). However, there are cases where curly brackets are always mandatory, enclosed in red curly brackets in the following examples:
\begin{itemize}
    \item In integral expressions, consider the following example: ``\$\textbackslash{}int\_0 $\hat{}$ 6\textcolor{red}{\{}f'(x)\textbackslash{},dx\textcolor{red}{\}}\$'' represents $\boxed{\int_0^6{f'(x)\,dx}}$.
    \item If the function's input contains brackets at the leftmost and rightmost positions or is a non-numerical string with more than one character, consider the following examples: ``\$\textbackslash{}ln\textcolor{red}{\{}(5a)\textcolor{red}{\}}\$'' represents $\boxed{\ln{(5a)}}$, ``\$e=\textbackslash{}cos\textcolor{red}{\{}(100\{\textbackslash{}pi\}t+\textbackslash{}pi)\textcolor{red}{\}}\char`\~(\textbackslash{}pu\{V\})\$'' represents $\boxed{e=\cos{(100{\pi}t+\pi)}~(V)}$.
\end{itemize}

Samples from our proposed dataset can be found in Appendix~\ref{sampledata}. Furthermore, detailed guidelines for utilizing our dataset in further research are provided\footnote{\url{https://huggingface.co/datasets/uitnlp/ViGEText_17to23}}.
\section{Experiments}

In this section, we present baseline LLMs (see Section \ref{baseline}) and their setups for evaluation (see Section \ref{setup}). Furthermore, the experimental results are described in Section \ref{results}.

\subsection{Baseline models}\label{baseline}
This section presents large language models for assessing their capacity in symbol binding. We explore whether higher MCSB ability leads to higher multiple-choice task accuracy. We evaluate five-shot model performance on two literature datasets (ViMMRC 1.0 \cite{kietnv2020} and ViMMRC 2.0 \cite{sonlt2023}) and our proposed dataset, which were all introduced in Section~\ref{dataset}. 

\begin{itemize}
    \item \textbf{BLOOMZ:} \citeauthor{muennighoff-etal-2023-crosslingual} employed MTF (Multilingual Task Fitting) to fine-tune pre-trained multilingual BLOOM and mT5 model families, resulting in adapted versions referred to as \textbf{BLOOMZ} and mT0. Their findings highlight that fine-tuning these large multilingual language models using English prompts for English tasks enables them to generalize effectively to non-English languages that are part of the pretraining corpus. Furthermore, when fine-tuned on multilingual tasks using English prompts, these models exhibit enhanced performance on English tasks and tasks involving non-English languages, achieving numerous SOTA results in zero-shot scenarios.

    \item \textbf{LLaMA:} \citeauthor{touvron2023llama} introduced LLaMA, a series of foundational language models spanning parameter counts from 7 billion to an impressive 70 billion. What makes LLaMA remarkable is its training on trillions of tokens, showcasing the possibility of training state-of-the-art models exclusively using publicly accessible datasets without the need for proprietary or inaccessible data sources.

    \item \textbf{GPT-3:} \citeauthor{brown2020language} trained \textbf{GPT-3}, an autoregressive language model with 175 billion parameters, 10x more than any previous non-sparse language model, and tested its performance in the few-shot setting. \textbf{GPT-3} is applied without any gradient updates or fine-tuning for all tasks, with tasks and few-shot demonstrations specified purely via text interaction with the model. \textbf{GPT-3} achieves strong performance on many NLP datasets.

    \item \textbf{GPT-4:} \textbf{GPT-4}, which was reported by \citeauthor{OpenAI2023GPT4TR}, is a large multimodal model that accepts text and image inputs and generates text outputs. \textbf{GPT-4} is a Transformer-based model trained to predict the next token in text documents. While it may not match human capabilities in all real-world situations, it excels on professional and academic benchmarks, including passing a simulated bar exam with a top 10\% score.
\end{itemize}

\begin{figure}[ht]
    \centering
    \includegraphics[width=\columnwidth]{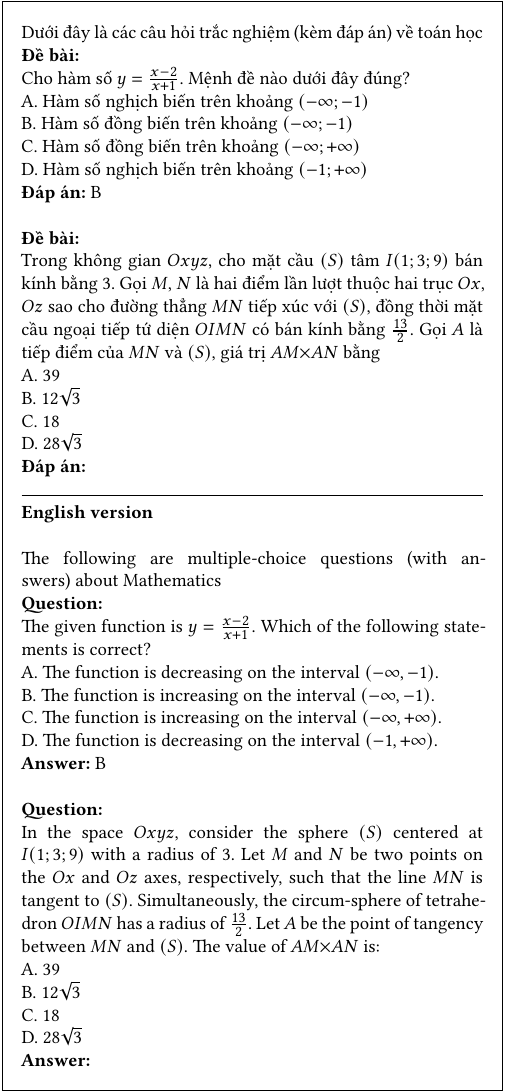}
    \caption{A mathematics example of one-shot learning of our proposed dataset. In this one-shot learning example, there is one instruction example and one initially incomplete example.}
    \label{prompt}
\end{figure}

For more detail, BLOOMZ-7.1B-MT, LLaMA-2-7B, LLaMA-2-70B, GPT-3, GPT-3.5, and GPT-4 are implemented for evaluating the symbol binding ability for multiple-choice questions in Vietnamese General Education. For LLaMA-2-7B and LLaMA-2-70B, we use the Replicate API\footnote{\url{https://replicate.com/}}. For GPT-3, GPT-3.5, and GPT-4, we use the OpenAI API\footnote{\url{https://openai.com/blog/openai-api}}. We deployed GPT-3.5 and GPT-4 in the May 12, 2023 version\footnote{These versions were released before the General Examination conducted; therefore, there is no leakage information in 2023.}. Finally, we perform inference on the BLOOMZ-7.1B-MT model with full precision using an NVIDIA A100 GPU provided by Google Colab.

In ViMMRC, our dataset, and multiple-choice prompts in general, we present a question and its answer choices as a single prompt to an LLM. The prompt is designed so that the model predicts only one token. The model's selected answer corresponds to the token with the highest probability. We treat the highest probability option as the prediction for each sample.

\subsection{Setup}\label{setup}

In this section, our setup for experiments is deputed.

\textbf{Few-shot prompt}
Following \citeauthor{hendryckstest2021}, we feed large language models (presented in Section~\ref{baseline}) prompts like that shown in Figure~\ref{prompt}. We begin each prompt with ``The following are multiple choice questions (with answers) about [subject].'' For zero-shot evaluation, we append the question to the prompt. For few-shot evaluation, we add up to 5 demonstration examples with answers to the prompt before appending the question. All prompts end with ``Answer: ''). The model then produces probabilities for the tokens ``A'', ``B'', ``C'', and ``D'' (ViMMRC 2.0~\cite{sonlt2023} does not constrain the questions to have fixed four options), and we treat the highest probability option as the prediction. To ensure consistent evaluation, we created a test set with 5 fixed few-shot examples from the \textbf{sample test} of 2017 published by the Vietnamese Ministry of Education for each subject.

\textbf{Evaluation Metric:} We initialize Accuracy for MCSB in this study. The formula is as follows in Equation~\ref{accuracy}:

\begin{equation}
    \text{Accuracy}=\frac{\text{Number of correct prediction tokens}}{\text{Total number of tokens}}\label{accuracy}
\end{equation}

\begin{figure}[ht]
    \centering
    \includegraphics[width=0.65\columnwidth]{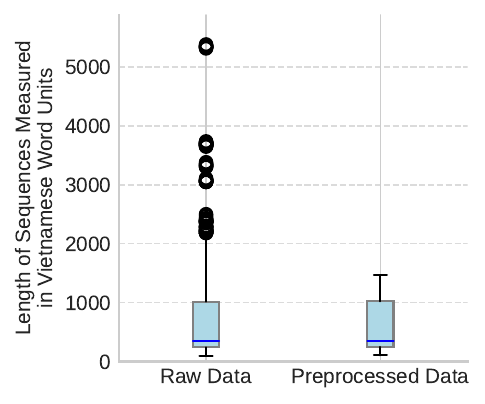}
    \caption{Distribution of sequence lengths, measured in Vietnamese word units using VnCoreNLP \cite{vu-etal-2018-vncorenlp}, for both raw data and preprocessed data, to ensure they do not exceed the maximum sequence length allowed by GPT-4.}
    \label{fig:x}
\end{figure}

\textbf{Max sequence length:} We established a maximum sequence length of $4096$ tokens for all large language models except for GPT-4, for which we specified a maximum sequence length of $3073$ tokens due to an unexcepted error from OpenAI API. ViMMRC 2.0 is a corpus focused on literature collected from literature schoolbooks, which included extremely long paraphrases (over $3073$ and $4096$ tokens). Therefore, to ensure that the sequence lengths of ViMMRC 2.0 do not exceed the max sequence length GPT-4 and other LLMs ($3073$ and $40961$ tokens, respectively), we implemented SentenceTransformer\footnote{\url{https://huggingface.co/sentence-transformers/paraphrase-xlm-r-multilingual-v1}} \cite{reimers-gurevych-2019-sentence} to rank the passage and remove unnecessary sentences. In Figure \ref{fig:x}, the lengths of sentences after and before ranking and removing have been presented for GPT-4 on ViMMRC 2.0, and this pattern is similarly observed for other LLMs.

\textbf{Max new tokens:} We have configured a maximum limit of $1$ token for generating responses to align with the constraints of the Multiple-Choice Questions task in Vietnamese General Education in this study.

\textbf{Temperature parameter:} The temperature parameter has been set to the value of $0$ to enhance result repeatability and facilitate reproducibility.

\textbf{Tokenizer:} For analysis in Figure~\ref{fig:x}, we implemented tiktoken\footnote{\url{https://github.com/openai/tiktoken}}, a high-speed tokenizer based on Byte-Pair Encoding (BPE), specifically designed to complement OpenAI's models.

\subsection{Results}\label{results}

\subsubsection{Experiments results on ViMMRC}

\begin{table}[ht]
\centering
\resizebox{\columnwidth}{!}{%
\begin{tabular}{|ll|c|c|}
\hline
\multicolumn{2}{|c|}{\textbf{Method}} & \textbf{ViMMRC 1.0} & \textbf{ViMMRC 2.0} \\ \hline\hline
\multicolumn{2}{|l|}{\textbf{Boosted score with ELMo} \cite{kietnv2020}} & 61.81 & \multicolumn{1}{c|}{--} \\ \hline
\multicolumn{2}{|l|}{\textbf{$\text{mBERT}_\text{cased}$} \cite{sonlt2021}} & 60.50 & \multicolumn{1}{c|}{--} \\ \hline
\multicolumn{2}{|l|}{\textbf{$\text{MMM}_\text{viBERT\&ViNLI}$} \cite{sonlt2023}} & 80.16 & 58.81 \\ \hline\hline
\multicolumn{1}{|l|}{\multirow{3}{*}{\textbf{BLOOMZ-7.1B-MT}}} & Zero-Shot & 79.96 & 64.47 \\
\multicolumn{1}{|l|}{} & One-Shot & 74.51 & 59.78 \\
\multicolumn{1}{|l|}{} & Five-Shot & 70.04 & 56.36 \\ \hline\hline
\multicolumn{1}{|l|}{\multirow{3}{*}{\textbf{LLaMA-2-7B}}} & Zero-Shot & 33.46 & 31.29 \\
\multicolumn{1}{|l|}{} & One-Shot & 40.47 & 35.35 \\
\multicolumn{1}{|l|}{} & Five-Shot & 39.69 & 36.70 \\ \hline\hline
\multicolumn{1}{|l|}{\multirow{3}{*}{\textbf{LLaMA-2-70B}}} & Zero-Shot & 75.10 & 64.83 \\
\multicolumn{1}{|l|}{} & One-Shot & 78.02 & 71.33 \\ 
\multicolumn{1}{|l|}{} & Five-Shot & 77.82 & 72.50 \\ \hline\hline
\multicolumn{1}{|l|}{\multirow{3}{*}{\textbf{GPT-3}}} & Zero-Shot & 77.04 & 68.53 \\
\multicolumn{1}{|l|}{} & One-Shot & 79.77 & 70.51 \\
\multicolumn{1}{|l|}{} & Five-Shot & 79.57 & 69.61 \\ \hline\hline
\multicolumn{1}{|l|}{\multirow{3}{*}{\textbf{GPT-3.5}}} & Zero-Shot & 82.88 & 73.49 \\
\multicolumn{1}{|l|}{} & One-Shot & 83.46 & 73.49 \\
\multicolumn{1}{|l|}{} & Five-Shot & 84.44 & 72.05 \\ \hline\hline
\multicolumn{1}{|l|}{\multirow{3}{*}{\textbf{GPT-4}}} & Zero-Shot & 90.86 & 84.22 \\
\multicolumn{1}{|l|}{} & One-Shot & 91.63 & 85.03 \\
\multicolumn{1}{|l|}{} & Five-Shot & 90.66 & 85.84 \\ \hline
\end{tabular}%
}
\caption{Experimental results of LLMs on ViMMRC 1.0 and ViMMRC 2.0 datasets.}
\label{tab:my-table-1}
\end{table}

Table~\ref{tab:my-table-1} deputed the results of previous works and large language models on ViMMRC 1.0 and ViMMRC 2.0 datasets. The results deputed that only GPT-3.5 and GPT-4 outperformed old SOTA models \cite{sonlt2023}, while most LLMs surpassed except for BLOOMZ-7.1B-MT and LLaMA-2-7B, the two least parameters LLMs. This claims an essential relationship between model architecture, scale, and performance on multiple-choice datasets. Moreover, this result also suggested the efficiency of LLMs on MQCA in the Vietnamese Literature task.

The results show that GPT series models perform better than others; the larger the parameters, the better GPT models are. GPT-4 achieved the highest accuracy and outperformed other LLMs on ViMMRC 1.0 and ViMMRC 2.0 datasets on three few-shot prompting scenarios. GPT-3 and GPT-3.5 also gained positive performances, while both LLMs surpassed others. \citeauthor{brown2020language} also observes that larger GPT-3 models perform better, though progress tends to be steadier.

LLaMA-2-7B and BLOOMZ-7.1B-MT are the smallest LLMs implemented in this study. However, the results of these two models are contradictory. While LLaMA-2-7B performs poorly (which has the worst performances on ViMMRC 1.0 and ViMMRC 2.0) according to model size and training, BLOOMZ-7.1B-MT shows the potential ability when outdoing LLaMA-2-70B and GPT-3 on zero-shot. Moreover, BLOOMZ-7.1B-MT has competitive results on one-shot compared to other LLMs (except for GPT-3.5 and GPT-4). However, it's noteworthy that BLOOMZ-7.1B-MT does not leverage the benefits of prompting. In our evaluations, we observed that BLOOMZ-7.1B-MT achieved its peak performance in the zero-shot scenario but experienced a decline in performance when transitioning to one-shot and five-shot scenarios. This observation underscores the distinct behavior of this model in contrast to others when prompted with varying levels of contextual information.

\begin{table*}[t]
\centering
\resizebox{\textwidth}{!}{%
\begin{tabular}{|ll|c|c|c|c|c|c|c|c|}
\hline
\multicolumn{2}{|c|}{\textbf{\begin{tabular}[c]{@{}c@{}}Large\\ Language Model\end{tabular}}} & \textbf{Mathematics} & \textbf{Physics} & \textbf{Chemistry} & \textbf{Biology} & \textbf{History} & \textbf{Geography} & \textbf{\begin{tabular}[c]{@{}c@{}}Civic\\ Education\end{tabular}} & \textbf{Average} \\ \hline\hline
\multicolumn{1}{|l|}{\multirow{3}{*}{\textbf{BLOOMZ-7.1B-MT}}} & Zero-Shot & 25.25 & 36.04 & 34.25 & 40.00 & 49.83 & 48.46 & 68.17 & 43.14 \\
\multicolumn{1}{|l|}{} & One-Shot & 22.77 & 30.57 & 30.12 & 35.05 & 43.33 & 24.62 & 59.17 & 35.09 \\
\multicolumn{1}{|l|}{} & Five-Shot & 25.25 & 28.68 & 32.19 & 31.74 & 40.33 & 43.46 & 64.33 & 38.00 \\ \hline\hline
\multicolumn{1}{|l|}{\multirow{3}{*}{\textbf{LLaMA-2-7B}}} & Zero-Shot & 24.59 & 23.96 & 28.57 & 26.61 & 28.83 & 32.31 & 27.33 & 27.46 \\
\multicolumn{1}{|l|}{} & One-Shot & 25.58 & 23.77 & 28.74 & 26.24 & 28.50 & 28.08 & 27.67 & 26.94 \\
\multicolumn{1}{|l|}{} & Five-Shot & 27.06 & 24.15 & 22.89 & 26.97 & 26.33 & 27.69 & 33.83 & 26.99 \\ \hline\hline
\multicolumn{1}{|l|}{\multirow{3}{*}{\textbf{LLaMA-2-70B}}} & Zero-Shot & 32.67 & 37.55 & 35.63 & 41.10 & 49.00 & 46.15 & 53.83 & 42.28 \\
\multicolumn{1}{|l|}{} & One-Shot & 35.31 & 42.83 & 37.87 & 37.43 & 52.00 & 41.54 & 67.50 & 44.93 \\
\multicolumn{1}{|l|}{} & Five-Shot & 34.16 & 40.57 & 36.14 & 43.30 & 55.67 & 41.92 & 67.67 & 45.63 \\ \hline\hline
\multicolumn{1}{|l|}{\multirow{3}{*}{\textbf{GPT-3}}} & Zero-Shot & 36.47 & 36.98 & 36.49 & 37.06 & 43.00 & 40.00 & 58.50 & 41.21 \\
\multicolumn{1}{|l|}{} & One-Shot & 40.76 & 38.30 & 41.14 & 42.20 & 43.50 & 40.38 & 63.17 & 44.21 \\
\multicolumn{1}{|l|}{} & Five-Shot & 40.10 & 39.43 & 41.48 & 43.12 & 47.83 & 41.54 & 67.33 & 45.83 \\ \hline\hline
\multicolumn{1}{|l|}{\multirow{3}{*}{\textbf{GPT-3.5}}} & Zero-Shot & 24.42 & 36.04 & 39.76 & 45.69 & 57.50 & 53.85 & 67.67 & 46.42 \\
\multicolumn{1}{|l|}{} & One-Shot & 38.94 & 43.96 & 49.91 & 50.28 & 57.50 & 51.54 & 69.67 & 51.69 \\
\multicolumn{1}{|l|}{} & Five-Shot & 40.26 & 44.72 & 50.09 & 51.38 & 60.17 & 51.54 & 72.67 & 52.97 \\ \hline\hline
\multicolumn{1}{|l|}{\multirow{3}{*}{\textbf{GPT-4}}} & Zero-Shot & 24.09 & 47.92 & 37.69 & 57.80 & 75.00 & 61.15 & 87.00 & 55.81 \\
\multicolumn{1}{|l|}{} & One-Shot & 55.45 & 66.04 & 53.36 & 64.77 & 78.50 & 68.46 & 88.50 & 67.87 \\
\multicolumn{1}{|l|}{} & Five-Shot & 56.44 & 66.60 & 64.20 & 69.17 & 82.17 & 71.92 & 88.17 & 71.24 \\ \hline
\end{tabular}%
}
\caption{Experimental results of LLMs on our proposed datsets.}
\label{tab:my-table-2}
\end{table*}

\subsubsection{Experiments results on our proposed dataset}

Table~\ref{tab:my-table-2} deputed the results of large language models on our proposed dataset. Unsurprisingly, GPT-4 achieved 55.81\%, 67.87\%, and 71.24\%  on average for our proposed dataset's zero-shot, one-shot, and five-shot scenarios and outperformed other LLMs. GPT series also obtained better performances than LLaMA and BLOOMZ. LLaMA-2-70B emerged as the second-highest performer in our evaluation, showcasing its commendable proficiency in symbol binding tasks. This finding suggested the huge efficiency of LLMs on MQCA in the Vietnamese General Education task.

In contrast, LLaMA-2-7B, while still a competent LLM, exhibited lower performance than its larger counterpart, LLaMA-2-70B. This discrepancy can be attributed to model size and training data differences. Smaller models often face limitations in capturing complex patterns and nuances, essential for symbol binding tasks. However, BLOOMZ-7.1B-MT, despite sharing a similar model size (7B parameters) with LLaMA-2-7B, achieved distinct results. It performed admirably with average accuracy rates of 43.14\%, 35.09\%, and 38.00\% for the few-shot settings.

Notably, in Mathematics, GPT-4 exhibited subpar performance in the zero-shot setting, achieving an accuracy rate of only 24.09\%, representing the lowest result among all the LLMs in our baseline. Following closely, GPT-3.5 achieved a slightly higher accuracy of 24.42\%, marking the second poorest performance in the zero-shot setting. In contrast, GPT-3 demonstrated impressive capabilities in the zero-shot setting. However, as we transitioned to the few-shot setting, we observed a notable improvement in GPT-3.5 and GPT-4 performance. This observation underscores the significance of the few-shot approach and its influence on GPT models and LLMs in a broader context, a topic we explore in greater detail in Section \ref{discussion}.
Moreover, in Geography, History, and Civic Education, it's important to note that current LLMs have not yet reached a perfect performance. They occasionally provide inaccurate answers on these subjects. There's a need for ongoing improvement to enhance their accuracy and effectiveness in these specific domains.

\section{Discussion}\label{discussion}

\begin{figure}[!ht]
    \centering
    \includegraphics[width=\columnwidth]{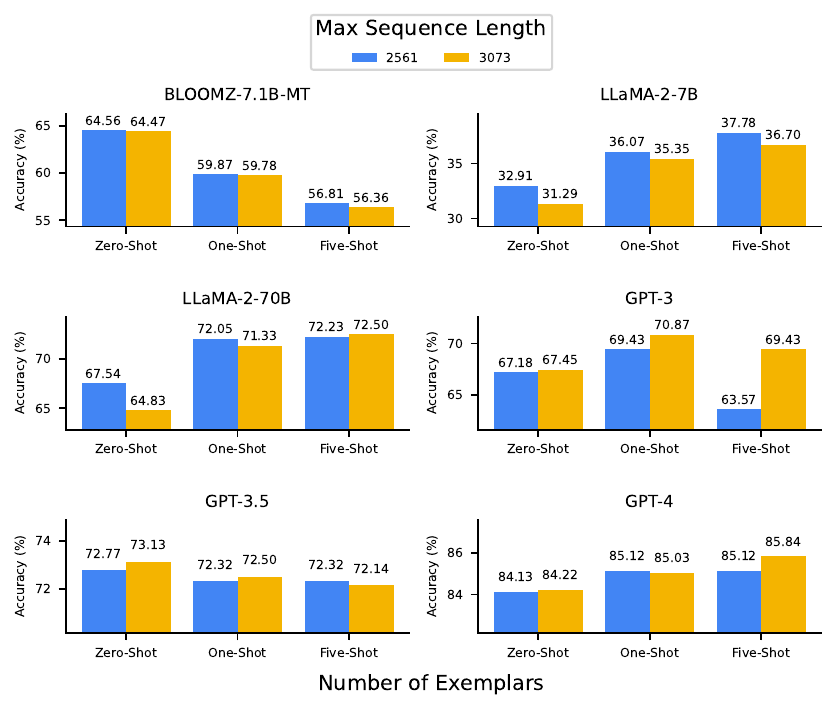}
    \caption{Performance scores on ViMMRC 2.0 for LLMs with varying maximum sequence lengths and numbers of exemplars.}
    \label{fig:22}
\end{figure}

\begin{figure*}[t]
    \centering
    \includegraphics[width=\textwidth]{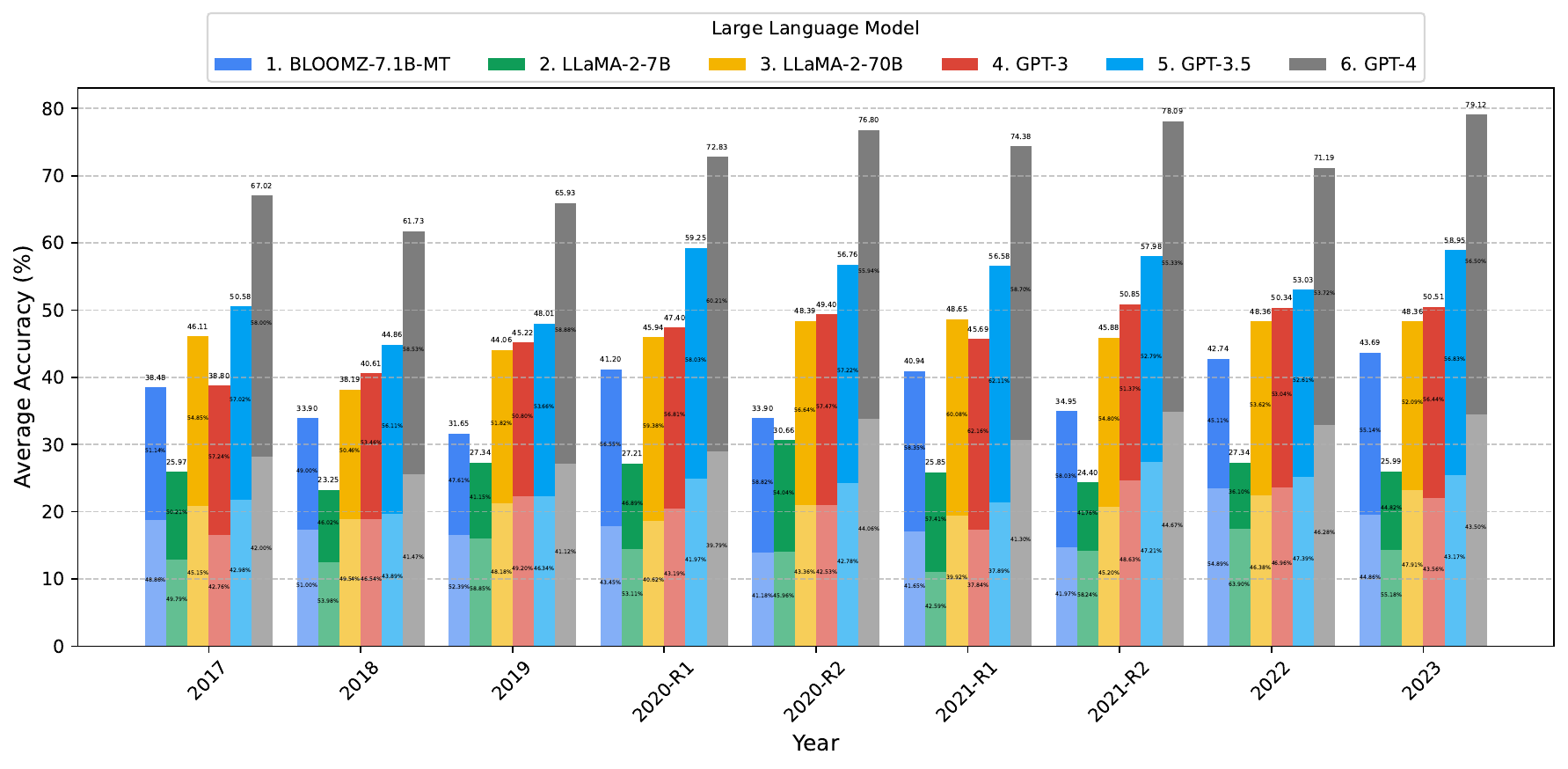}
    \caption{Performance average scores of LLMs on our proposed dataset from 2017 to 2023. The bottom of each column with a lighter shade denotes the second half of every test, as it is consistently more challenging than the initial half, as per the Vietnamese Ministry of Education.}
    \label{fig:accbyyear}
\end{figure*}

As observed in Figure \ref{fig:22}, the GPT model series demonstrates improved performance when subjected to longer maximum sequence lengths. Conversely, other Large Language Models (LLMs) tend to yield superior results when constrained by shorter maximum sequence lengths. Furthermore, it is worth highlighting a distinct finding: the BLOOMZ model exhibits a notably diminished performance when prompted. In summary, this finding underscores the significant impact of both maximum sequence length and the number of prompts on the performance of LLMs within the specific context of the dataset used in this research. The right parameters are crucial for effective models, highlighting the need for tailored tuning in LLM applications, especially in education and beyond.

Figure~\ref{fig:accbyyear} presented the average accuracy of each LLM on our proposed dataset for each year (from 2017 to 2023) on five-shot setting. The results show that LLMs struggled with the Vietnamese General Education Examination in 2017 and 2018. However, there has been a noteworthy shift in LLMs' performance from 2020 to 2023. The primary reason lies in the evolving nature of the examinations themselves. Overall, GPT-3.5 and GPT-4 consistently demonstrated the most impressive performances.

The examinations conducted in 2017 and 2018 were acknowledged as the most challenging general education assessments. They featured complex and demanding questions that posed a formidable challenge for LLMs. Conversely, in more recent years, starting from 2020, the examinations were intentionally made less difficult, with questions designed to be easier and less intricate. Moreover, among the examinations, GPT-4 and other LLMs solved mostly 60\% on the first exam questions, while these LLMs struggled with other 40\%. This is because mostly 60\% of the first exam questions are much easier than the 40\% less. This finding highlights the critical role of examination difficulty in LLMs' performances.

\section{Conclusion and Future Work}
In this study, we investigated the multiple-choice symbol binding (MCSB) abilities of large language models (LLMs) in Vietnamese multiple-choice question answering (MCQA). Our contributions included the creation of a novel, high-quality dataset, the rigorous evaluation of six prominent LLMs, and an in-depth analysis of their impact on Vietnamese MCQA, especially in General Education.

Our novel dataset enforces strict LaTeX guidelines, ensuring easy parsing in future research. Designed for MCQA in subjects like Mathematics, Physics, Chemistry, and Biology, it fills a vital gap for Vietnamese. This standardized resource facilitates comprehensive assessments across diverse domains for LLMs.

We extensively tested BLOOMZ-7.1B-MT, LLaMA-2-7B, LLaMA-2-70B, GPT-3, GPT-3.5, and GPT-4.0 in Vietnamese MCQA tasks. Our evaluations highlighted their strengths and weaknesses, deepening our understanding of their capabilities.

\section*{Limitations}
This paper offers an evaluation without detailed analysis or explanation, pointing toward future research. Reproducing GPT-X results is challenging due to limited open-source access. However, all experiment results are included\footnote{\url{https://github.com/uitnlp/vigetext_17to23}}, aiding future research efforts.

\begin{acks}
This research was supported by The VNUHCM-University of Information Technology’s Scientific Research Support Fund. We are grateful to the reviewers for their valuable comments, which have greatly enhanced the quality of our work.
\end{acks}

\balance
\bibliographystyle{ACM-Reference-Format}
\bibliography{main}

\appendix

\section{Zero-Shot Prompts for Our Dataset}
\label{sampledata}

\begin{figure}[H]
    \centering
    \includegraphics[width=\columnwidth]{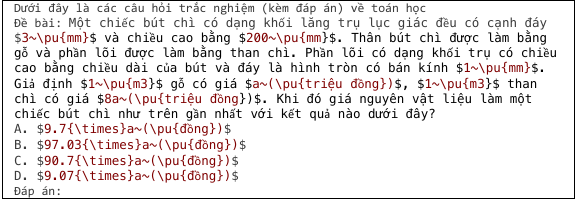}
    \caption{A Mathematics example.}
\end{figure}

\begin{figure}[H]
    \centering
    \includegraphics[width=\columnwidth]{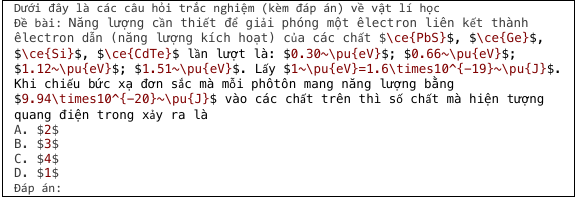}
    \caption{A Physics example.}
\end{figure}

\begin{figure}[H]
    \centering
    \includegraphics[width=\columnwidth]{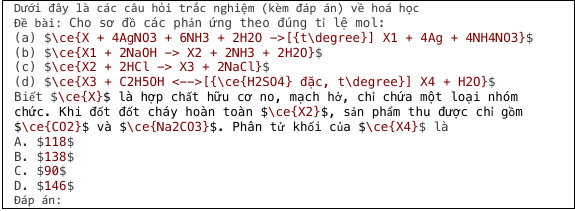}
    \caption{A Chemistry example.}
\end{figure}

\begin{figure}[H]
    \centering
    \includegraphics[width=\columnwidth]{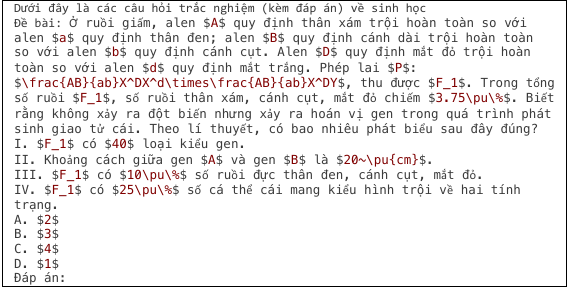}
    \caption{A Biology example.}
\end{figure}

\end{document}